\documentclass[10pt,journal,compsoc,anonymous]{IEEEtran}
\IEEEoverridecommandlockouts
\usepackage{cite}
\usepackage{amsmath,amssymb,amsfonts}
\usepackage{algorithmic}
\usepackage{graphicx}
\usepackage{textcomp}
\usepackage{multirow}
\usepackage{xcolor}
\def\BibTeX{{\rm B\kern-.05em{\sc i\kern-.025em b}\kern-.08em
    T\kern-.1667em\lower.7ex\hbox{E}\kern-.125emX}}
\begin{document}

\title{Epidemic Decision-making System Based Federated Reinforcement Learning}

\author{Yangxi Zhou, 
        Junping Du*,
        Zhe Xue,
        Zhenhui Pan, 
        and Weikang Chen
\IEEEcompsocitemizethanks{\IEEEcompsocthanksitem Yangxi Zhou, Junping Du, Zhe Xue, Zhenhui Pan, and Weikang Chen are with the School of Computer Science (National Pilot School of Software Engineering), Beijing University of Posts and Telecommunications, Beijing Key Laboratory of Intelligent Telecommunication Software and Multimedia, Beijing 100876, PR China. 
E-mail: zhouyx@bupt.edu.cn, junpingdu@126.com, xuezhe@bupt.edu.cn, JerryPan@bupt.edu.cn, weikangchen@bupt.edu.cn.
}

\thanks{*Corresponding author: Junping Du (junpingdu@126.com).}
\thanks{This work was supported by the Program of the National Natural Science Foundation of China (62192784, U22B2038, 62172056).}
}

\markboth{The 5-th International Conference on Data-driven Optimization of Complex Systems}%
{Zhou \MakeLowercase{\textit{et al.}}: Epidemic Decision-making System Based Federated Reinforcement Learning}

\IEEEtitleabstractindextext{%
\begin{abstract}
Epidemic decision-making can effectively help the government to comprehensively consider public security and economic development to respond to public health and safety emergencies. Epidemic decision-making can effectively help the government to comprehensively consider public security and economic development to respond to public health and safety emergencies. Some studies have shown that intensive learning can effectively help the government to make epidemic decision, thus achieving the balance between health security and economic development. Some studies have shown that intensive learning can effectively help the government to make epidemic decision, thus achieving the balance between health security and economic development. However, epidemic data often has the characteristics of limited samples and high privacy. However, epidemic data often has the characteristics of limited samples and high privacy. This model can combine the epidemic situation data of various provinces for cooperative training to use as an enhanced learning model for epidemic situation decision, while protecting the privacy of data. The experiment shows that the enhanced federated learning can obtain more optimized performance and return than the enhanced learning, and the enhanced federated learning can also accelerate the training convergence speed of the training model. accelerate the training convergence speed of the client. At the same time, through the experimental comparison, A2C is the most suitable reinforcement learning model for the epidemic situation decision-making. learning model for the epidemic situation decision-making scenario, followed by the PPO model, and the performance of DDPG is unsatisfactory.
\end{abstract}

\begin{IEEEkeywords}
Federated learning, reinforcement learning, decision-making, epidemic
\end{IEEEkeywords}}

\maketitle

\section{Introduction}
With the outbreak of New Crown Pneumonia in 2020, the epidemic has had a tremendous impact on the safety of life and property in human society. Due to the complexity and uncertainty of the outbreak situation, the development of outbreak policies needs to be considered comprehensively in order to find a balance between life and health and economic development, which poses a variety of challenges to policy makers. In past outbreaks, there has been a lack of emergency preparedness for public health and safety emergencies\cite{b1}. For this reason, recent research has proposed a reinforcement learning-based decision-making system for outbreaks, however, outbreak data is often difficult to collect, has small samples, and is highly private, which prevents individual decision makers from sharing data and thus cooperating with each other. Federated learning is a machine learning method that can combine datasets from different locations to learn while protecting data privacy and security\cite{b2}.

In order to solve the problems of small samples and privacy of epidemic data, this paper introduces federated learning based on reinforcement learning for epidemic decision-making system, which unites the epidemic data of each province without disclosing the privacy, and promotes the decision makers to make better decisions with the support of the epidemic decision-making system.
The main contribution of this paper consists of 3 areas.

1) we propose a reinforcement federated learning-based epidemic decision-making system that enables better epidemic decision recommendations for decision makers in different provinces while protecting data privacy and security;

2) Analyze the performance of different reinforcement learning models in epidemic decision-making scenarios on the basis of federated learning, and summarize the training effect of different reinforcement learning models;

3) Analyze and compare the trajectory of model training when using reinforcement federation learning and reinforcement learning. Reinforcement federation learning is found to result in faster model training convergence and higher cumulative returns. 

\section{Related Work}
In this section, we provide a brief overview of the current state of research on federated learning, reinforcement learning, federated reinforcement learning, and epidemic decision making.
\subsection{Federated Learning}
With the demand for privacy, many countries have introduced regulations to protect privacy, and the approach of centralizing private data from different places and then learning from it will meet a lot of obstacles. Federated learning is a method that can iteratively aggregate local model updates from multiple clients to obtain a global model without compromising data privacy.\cite{b3,b4,b5,b6,b7,b8} Federal Learning. Federated learning solves the data privacy problem in big data scenarios.McMahan et al. identified the problem of training decentralized data from mobile devices as an important research direction i.e. federated learning and proposed FedAvg, a deep network federated learning method for model averaging through iterations, to validate that the proposed method has high performance\cite{b2}.
\subsection{Reinforcement Learning}
Reinforcement learning allows an intelligent body to interact with the environment it is in, the intelligent body makes an action on the environment based on the observed state of the environment, the environment switches states based on the action of the intelligent body and gives the intelligent body a reward, and the learning goal of reinforcement learning is to allow the intelligent body to obtain the highest cumulative reward possible\cite{b9,b10,b11,b12,b13,b14,b15}. The goal of reinforcement learning is to enable intelligences to achieve the highest cumulative returns. In the past decade, reinforcement learning has achieved significant success in decision-making problems, such as AlphaGo. Recent reinforcement learning has gained stronger performance by incorporating deep neural networks and interacting with dynamic environments.Schulman et al.\cite{b9} proposed a proximal policy optimization PPO algorithm that alternates between sampling data from a policy and performing multiple rounds of optimization on the sampled data, which is simpler to implement, more general, and has better sample complexity.A2C\cite{b10} Proposed by OpenAI as an alternative to asynchronous implementations, researchers have found that it is possible to write a synchronous, deterministic implementation that waits for each actor to complete its experience fragment before performing an update that averages across all actors. An advantage of this approach is that it makes more efficient use of the GPU and performs best at high volumes. Deep Deterministic Policy Gradient (DDPG) \cite{b11} is an algorithm that learns both Q-functions and policies. It uses off-policy data and Bellman's equation to learn the Q-function and uses the Q-function to learn the policy.Twin Delayed Deep Deterministic policy gradient algorithm (TD3)\cite{b12} is an actor-critic algorithm that considers the interaction between the function approximation error in policy and value updates.
\subsection{Federated Reinforcement Learning}
Although reinforcement learning and deep reinforcement learning have excelled in many areas, they still face several important technical and non-technical challenges in solving real-world problems\cite{b16,b17,b18,b19,b20,b21,b22,b23}. The successful application of federation learning in supervised learning tasks has generated interest in using a similar idea in reinforcement learning, namely reinforcement federation learning. Although federated learning is useful in some specific situations, it cannot handle collaborative control and optimal decision making in dynamic environments. Reinforcement federated learning\cite{b16} not only provides experience for intelligences to learn to make correct decisions in unknown environments, but also ensures that the data collected during the intelligences' exploration does not have to be shared with others.

Reinforcement federation learning can be categorized into horizontal reinforcement federation learning and vertical reinforcement federation learning. In horizontal reinforcement federation learning, the environment in which each intelligent body interacts is independent of the other environments, while the state spaces and action spaces of different intelligences are aligned to solve similar problems. Each intelligent's behavior affects only its own environment and produces a corresponding reward. In longitudinal reinforcement federated learning, multiple intelligences interact with the same global environment, but each intelligence can only observe a limited amount of state information within its visible range. Intelligents can perform different operations based on the observed environment and receive local rewards or even no rewards. Since the epidemic environments between different provinces in the epidemic scenario are relatively independent, and the decision-making is the responsibility of each department separately, this paper adopts horizontal reinforcement federation learning, where the intelligences in each province interact in independent environments.Jin et al.\cite{b17} proposed reinforcement federation learning algorithms QAvg and PAvg based on horizontal reinforcement federation learning, and theoretically proved that these algorithms converge to a suboptimal solution, and this suboptimality depends on the heterogeneity of these n environments.
\subsection{Epidemic Decision-making}
As the new coronary pneumonia continues to infect the world, the speed and scope are unprecedented. How to make reasonable decisions in time to ensure the public health safety of people to the maximum extent while ensuring the smooth economy is a problem of research value. In reality, to make a rational decision on epidemic needs to consider various factors, such as population density, epidemic transmission rate, incubation period, medical conditions, economic development and so on. Therefore, many researchers have investigated the use of reinforcement learning for epidemic decision making to support decision makers. At the same time, the small sample size of a single province for epidemic decision-making is also one of the problems\cite{b24,b25,b26,b27,b28,b29}.

Giordano et al.\cite{b30} proposed a new model to predict the course of an outbreak to help plan an effective control strategy. The model considers eight stages of infection: susceptible (S), asymptomatic infection unrecognized (I), asymptomatic infection recognized (D), symptomatic infection unrecognized (A), symptomatic infection recognized (R), severe (T), cured (H), and death (E), collectively referred to as SIDARTHE.The SIDARTHE model distinguishes between infected individuals based on whether they are recognized and the severity of their symptoms. them. They compared the simulation results with real data from the Italian COVID-19 outbreak and modeled possible scenarios for implementing countermeasures. Our results suggest that social distance restriction measures need to be combined with extensive nucleic acid detection and close contact tracing to end the ongoing neo-crest epidemic.Kompellad et al.\cite{b31} propose a new intelligentsia-based outbreak simulation environment that models fine-grained interactions between people in specific locations in a community; and propose a reinforcement learning-based approach for optimizing outbreak decisions.Kwak et al.\cite{b32} used deep reinforcement learning to try to find optimal public health strategies to maximize the total payoff of strategies to control the spread of new crowns. Learned intelligence will be implemented earlier than actual decision making, while being more intense, but does not advocate rapid total containment and closure of border gates.Padmanabhan et al.\cite{b33} Developed and used a closed-loop control strategy based on reinforcement learning as a decision-support tool for responding to a new crown outbreak, proposing a new model of disease transmission that considers the impact of non-pharmacological interventions on overall disease transmission rates and infection rates in both asymptomatic and symptomatic phases, as well as taking into account the characteristics of the outbreak, health-care system parameters, and socioeconomic factors.Zong et al.\cite{b34} proposed an intelligent-based multi-intelligent reinforcement learning neo-crown outbreak simulation environment based on Kompellad et al. which can simulate not only fine-grained interactions between people in a specific location, but also population movement between different U.S. states with different economic structures and age distributions, and a new multi-intelligent reinforcement learning algorithm capable of capturing the temporal interval characteristics of the environment. The results show that the algorithm has better performance and also calibrates the environment using real epidemic spread data to make the simulation results more consistent with the real situation.Chadi et al.\cite{b35} investigated a reinforcement learning based approach for automatically analyzing and recommending control policies under different constraints in an epidemic context and introduced a new epidemiological model (EM) based on the SIDARTHE model, which is suitable for analyzing different control policies and conducted experiments in ten cities, suggesting future directions that can be done with multi-intelligent agent reinforcement learning and viral heterogeneity.

In this paper, we propose an epidemic decision-making system based on reinforcement federated learning, which solves the problems of small sample size of a single province, the privacy issue that the model can only assist decision-making within one's own province, and the privacy issue of centralized training in multiple provinces.

\section{Federated reinforcement Learning Based Epidemic Decision-making System}
In this section, we focus on the reinforcement federation learning mechanism, j propose four specific reinforcement federation learning models (PPOAvg, A2CAvg, DDPGAvg, TD3Avg) based on QAvg, PAvg and apply them to epidemic decision-making scenarios.

\subsection{federated reinforcement learning Mechanisms}
\begin{figure}[htbp]
\centerline{\includegraphics[scale=0.7]{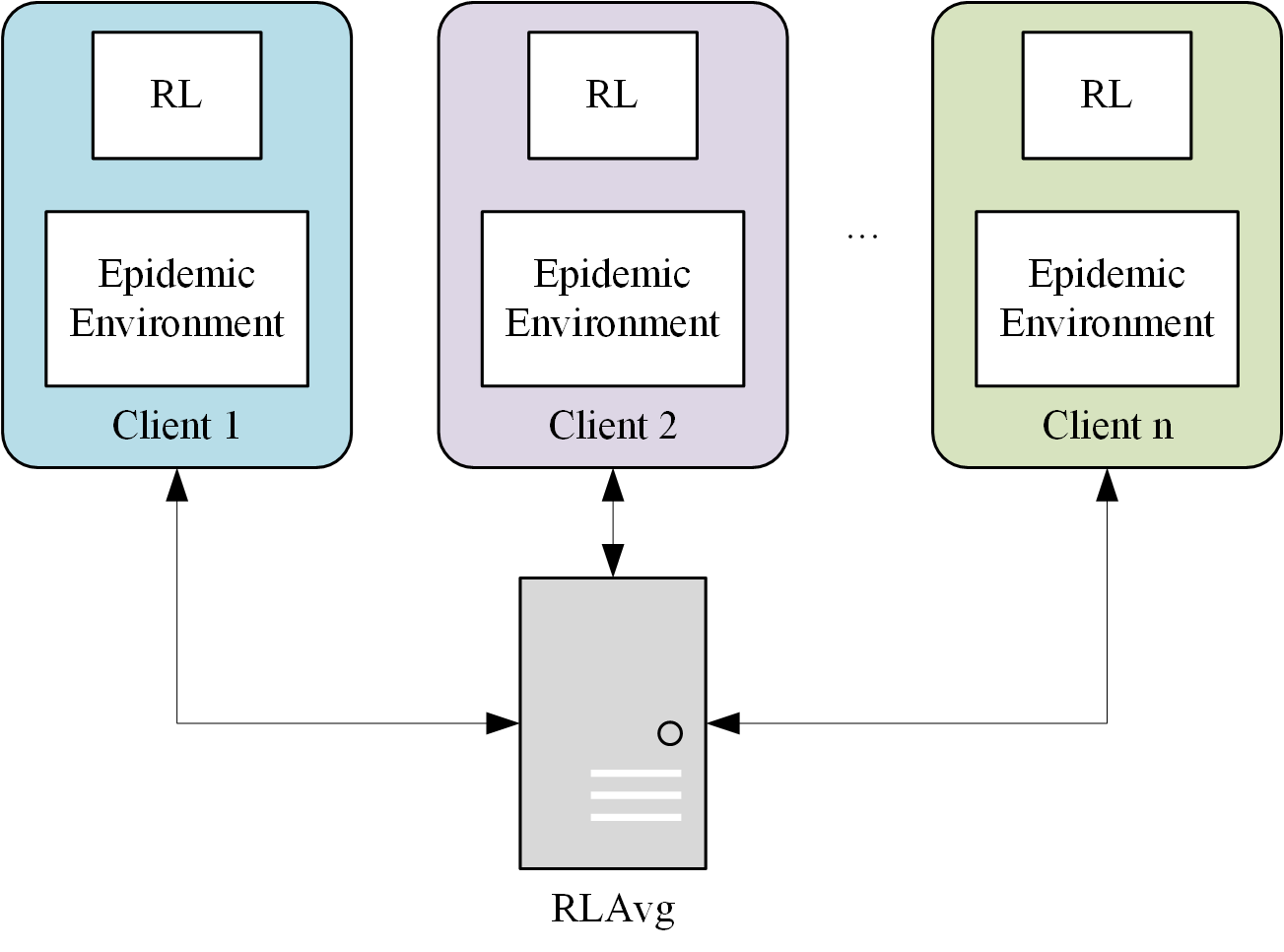}}
\caption{Illustration of RLAvg.}
\label{fig1}
\end{figure}

The classical federated learning algorithm FedAvg aggregates the model parameters of the clients thus obtaining the global model. Reinforcement federation learning, on the other hand, obtains the global model by aggregating the model parameters of the reinforcement learning model. Reinforcement learning can be categorized into policy-based reinforcement learning and value-based reinforcement learning, which correspond to PAvg and QAvg in reinforcement federation learning, respectively, and these algorithms alternate between local computation and global aggregation. Specifically, each client's intelligence updates its value function or policy function locally multiple times, and then the server averages the aggregation of these n functions across all agents. To improve communication efficiency, multiple local updates are performed between communication aggregations. Taking the PPO model as an example, the reinforcement federated learning mechanism PPOAvg is shown in Fig.~\ref{fig1}.

The specific steps are as follows:

1. Random selection: k clients are randomly selected for local training, each training local epochs round;

2. local update: Calculate the parameter updates of the trained local model, perform averaging and then update the global model to obtain a new global model;

3. Third, global aggregation: pass the new global model parameters to the selected client and then repeat step one, cycling through the updates until the completion of the global epochs round.

Where for QAvg, the only thing that is shared by the clients in the global aggregation phase is the Q-table, and not their collected experience; similarly, PAvg will only share the client's policy function during the whole training process. Following the above principles, this paper also implements a reinforcement federation learning model for different reinforcement learning algorithms of A2CAvg, DDPGAvg, and TD3Avg. The algorithms are trained in K parallel environments to improve the sampling efficiency and reduce the variance of updates.

\subsection{Epidemic Decision-making Environment}
Since the SIDARTHE epidemiological model is very effective for outbreak prediction, the outbreak decision-making environment in this paper uses the SIDARTHE model to simulate the real environment.
The environment defines seven decision-making actions and four state attributes, and the intelligences can choose different action intensity distributions according to different states to be rewarded for interacting with the environment. As shown in Tab.~\ref{tab1}, actions A1-A7 represent public health intervention policies, and states s1-s4 are used to portray the development of the current epidemic.
s1 transmission rate indicates the proportion of the normal population that will be infected, which is calculated using data such as A1, A2, A3, A4, A7, as well as population density and the number of people currently infected, as described in the work of Chadi et al.\cite{b35} 's work; s2 represents the probability of identifying a positive, which is related to the incubation period and nucleic acid detection rate, and can therefore be calculated as follows:

\begin{figure*}[htbp]
\centerline{\includegraphics{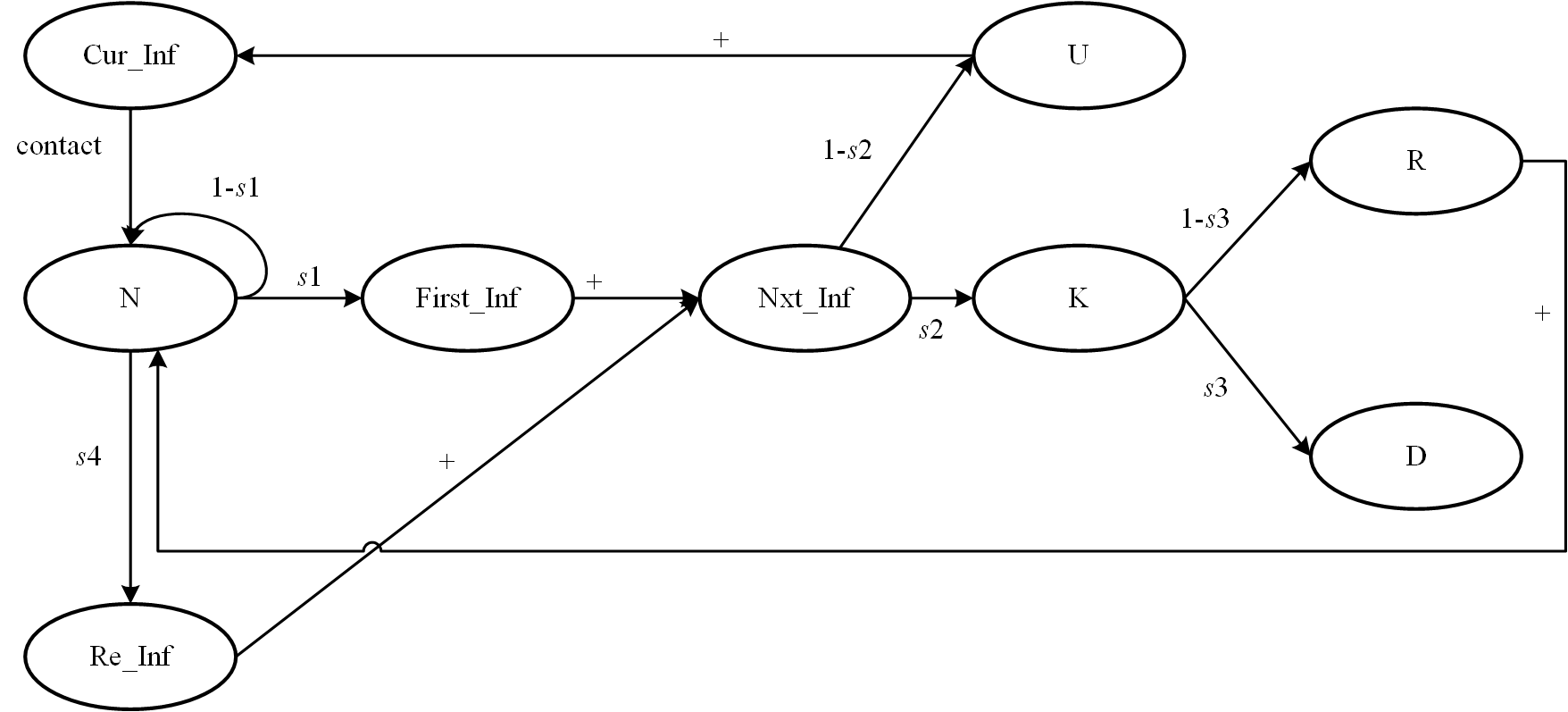}}
\caption{State Transition Model of Environment.}
\label{fig2}
\end{figure*}

\begin{equation}
    s2 = A5*(1 - \textit{incubation period})
\end{equation}

s3 represents the mortality rate, which is proportional to the inherent morbidity and mortality of the virus, and the greater the medical capability, the more it reduces the mortality rate, calculated as follows:

\begin{equation}
    s3 = (1 - A6) * \textit{fatality rate}
\end{equation}

s4 denotes the reinfection rate, which was defined as the proportion of invalidly vaccinated individuals (61$\%$) multiplied by the probability of reinfection (approximately 16$\%$):

\begin{equation}
    s4 = \left( 1 - (A7*0.61) \right) * \textit{Probability of reinfection}
\end{equation}

Based on the above epidemic dynamics modeling, the single iteration environmental state transition model can be shown in Fig.~\ref{fig2}, where \text{Cur\_Inf} is the current number of infections, N is the normal population, \text{First\_Inf} is the number of people infected for the first time, \text{Re\_Inf} is the number of people re-infected after recovering, \text{Nxt\_Inf} is the number of people who will be infected in the next round, U: Undiscovered carriers of the virus, K: Known virus carriers, R: number of recovered people, D: number of dead people. We consider that a region with a normal (i.e., uninfected) population "N" will be exposed to an initial number of "\text{Cur\_Inf}" currently infected people (e.g., from another infected region), which will produce s1*N first infections "\text{First\_Inf}" (i.e., no history of new crowns), and s4*R reinfected individuals "\text{Re\_Inf}", with R representing the number of recoveries. Both the first-infected and re-infected populations will be added to the next round of infected populations "\text{Nxt\_Inf}", which will be categorized into two other categories based on the recognition rate s2, i.e., known carriers "K" or unknown carriers "U ". Unknown carriers "U" (e.g., those who are asymptomatic or undetected) will be added to the currently infected population "\text{Cur\_Inf}" because "U" is not aware of U" is unaware of their infection and adds to the number of infected people. Known carriers (e.g., carriers who test positive and are hospitalized) are either converted into a dead population based on the mortality rate s3 and subtracted from the global population "N", or recovered and added to the global population. The formula for each population can be summarized as follows:

\begin{equation}
    \text{Cur\_Inf} = Cur{\_ I}nf + U
\end{equation}

\begin{equation}   
N = N - D
\end{equation}

\begin{equation}
    \text{First\_Inf} = s1*N
\end{equation}

\begin{equation}
\text{RI} = s4*R
\end{equation}

\begin{equation}
\text{RI} = s4*R
\end{equation}

\begin{equation}
    \text{Nxt\_Inf} = \text{First\_Inf} + \text{Re\_Inf}
\end{equation}

\begin{equation}
    K = s2*\text{Nxt\_Inf}
\end{equation}

\begin{equation}
    U = \left( {1 - s2} \right)*\text{Nxt\_Inf}
\end{equation}

\begin{equation}
    R = \left( {1 - s3} \right)*K
\end{equation}

\begin{equation}
    D = s3*K
\end{equation}

Since there are various objective and subjective factors in the implementation of epidemic policies, and there will always be variability in the actual implementation of each unit, the environment introduces randomness in determining the intensity of the action: when the government implements a strict and high-intensity policy, the intensity of the action is taken as a random number between 0.75-1; when a stricter-intensity policy is implemented, the intensity of the action is taken as a random number between 0.5-0.75 When a weaker policy is in place, the strength of the action is a random number between 0.2 and 0.5; when no policy is in place, the strength of the action is a random number between 0 and 0.2.
In reinforcement learning, rewards are used to measure the performance of an intelligent body. In an epidemic setting, this paper allows an intelligent body to find the best policy among seven actions, determining a series of actions to reduce the infection rate at the least cost to maintain economic development. For this purpose, the reward function is defined as:

\begin{equation}
    reward = \left\{ \begin{matrix}
{h + e,if~h > 0~and~e > 0} \\
{0,~others} \\
\end{matrix} \right.
\end{equation}

included among these

\begin{equation}
    h = \left\{ \begin{matrix}
    {{\left( {TI - Nxt{\_ I}nf} \right)/{TI}},if~Nxt{\_ I}nf < TI,~D < TD} \\
    {0,others} \\
    \end{matrix} \right.
\end{equation}

where TI and TD are thresholds that cannot be exceeded by the next round of infected and dead populations.

\begin{equation}
    e = 1 - \frac{\sum{wi*Ai}}{\sum{wi}}
\end{equation}

On the one hand, the health score h reflects the intelligence's performance in terms of public health security: the fewer infections than a set threshold TI, the greater h becomes, and once the number of infections exceeds TI or the number of deaths exceeds TD, h goes straight to 0. As Verschuur et al.\cite{b36} 's conclusions, coercive measures like prolonged embargoes can have serious consequences for the global economy. Therefore, the weights w1-w7 applied to actions (A1-A7) in the economic score are used to regulate the priority of performing some actions to guide the learning of the intelligence. That is, if the region where the intelligent is located has a policy that has a much higher impact on the economy than other policies, its weight should be much higher, and a small change in a highly weighted action will have a much greater impact on the score e. In addition, the scores of these two definitions are inversely proportional, i.e., to increase the economic score, intelligences should take no (or little) action, whereas to increase the health score, agents should apply high-intensity public health and safety policies (actions) to reduce the number of infections.

\begin{table}[htbp]
\caption{The Epidemic Decision-making Action of the Agent}
\begin{center}
\begin{tabular}{|c|c|}
\hline
\textbf{Action}&\textbf{Description} \\

\hline
A1 & Travel restriction \\
\hline

A2 & Lockdown \\
\hline

A3 & Distance work and education \\
\hline

A4 & Provide masks and impose their wearing \\
\hline
A5 & Increase the testing rate (test and isolate if positive) \\
\hline
A6 & Increase the health-care capacity (e.g., hospital beds) \\
\hline
A7 &  Increase the vaccination rate \\
\hline
S1 &  The transmission rate \\
\hline
S2 &  The identification rate \\
\hline
S3 &  The death rate \\
\hline
S4 &  The probability of reinfection \\
\hline

\end{tabular}
\label{tab1}
\end{center}
\end{table}

Since there are various objective and subjective factors in the implementation of epidemic policies, and there will always be variability in the actual implementation of each unit, the environment introduces randomness in determining the intensity of the action: when the government implements a strict and high-intensity policy, the intensity of the action is taken as a random number between 0.75-1; when a stricter-intensity policy is implemented, the intensity of the action is taken as a random number between 0.5-0.75 When a weaker policy is in place, the strength of the action is a random number between 0.2 and 0.5; when no policy is in place, the strength of the action is a random number between 0 and 0.2.

In reinforcement learning, rewards are used to measure the performance of an intelligent body. In an epidemic setting, this paper allows an intelligent body to find the best policy among seven actions, determining a series of actions to reduce the infection rate at the least cost to maintain economic development. For this purpose, the reward function is defined as:
\section{experiments}
\subsection{Experiment Settings}
Based on the above algorithms, this section provides experimental observations on the performance of reinforcement federation learning algorithms in epidemic decision-making scenarios. First, different reinforcement learning models are trained with reinforcement federation learning, while the validation rewards of the global model obtained by reinforcement federation learning are compared with those of the central model using a centralized training method. Second, the rewards of each client during the training process were compared using the PPO model as an example. For the parameter setting of federated learning, this paper sets up 10 clients for training, and 5 clients (k) are selected in each round for 3 rounds of local training (local epochs), and then aggregated to complete 20 rounds in total (global epochs). For the environmental epidemic parameter settings, this paper sets the incubation period to 14 days, the case fatality rate to 0.02, the population to 100,000, the initial infection to 3, the total population density to 1,000 people per square kilometer,and the infection threshold TI to 25.

\begin{figure*}[htbp]
\centerline{\includegraphics{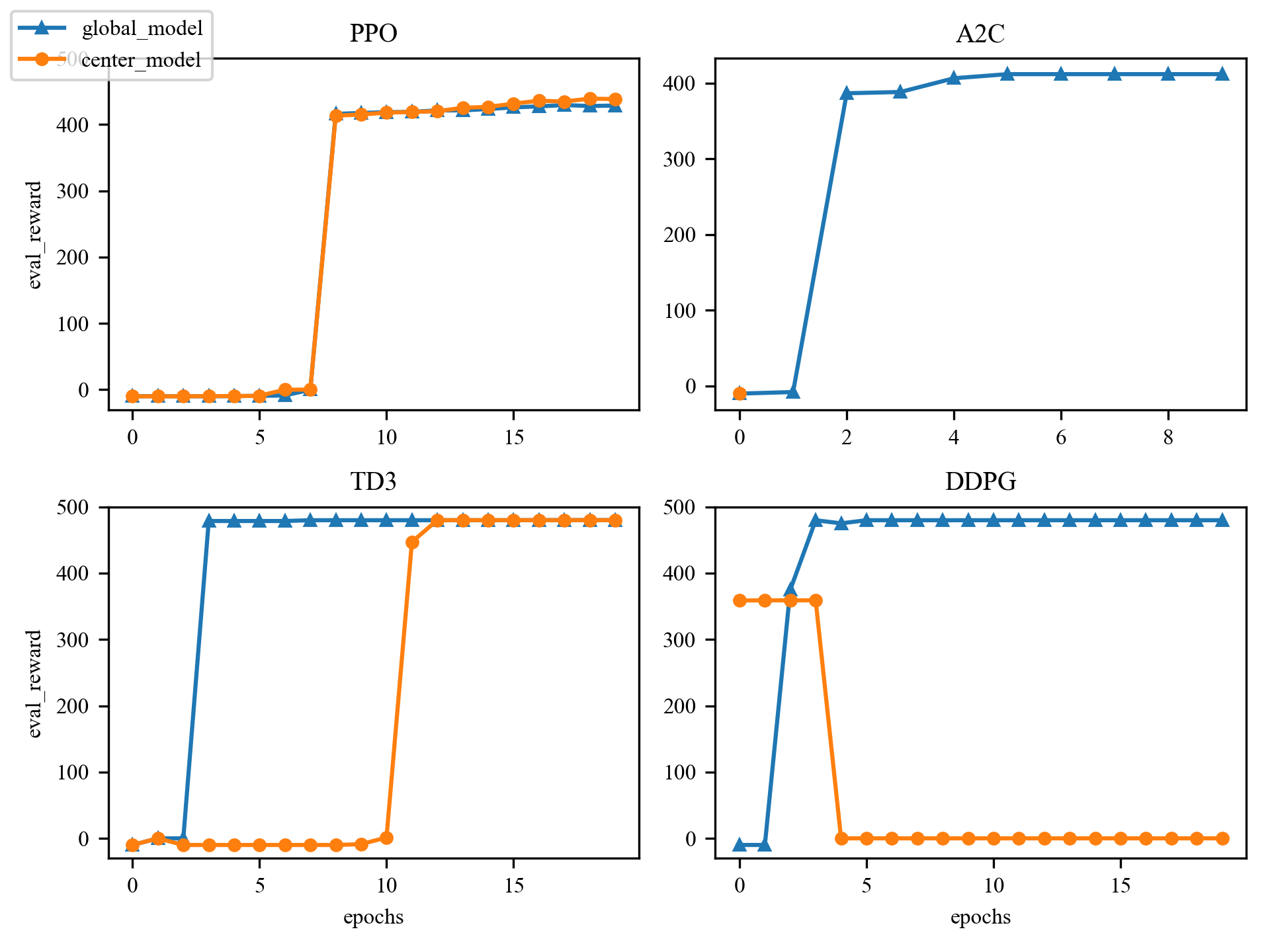}}
\caption{Diagram of Different Models Experimental Evaluation Results.}
\label{fig3}
\end{figure*}

\subsection{Results}
As shown in Fig.~\ref{fig3} is the validation rewards of reinforcement federated learning training using different reinforcement learning models, this paper mainly selected the mainstream reinforcement learning models such as PPO, DDPG, TD3, A2C, etc., and conducted experiments, which showed that the cumulative rewards of PPO's global model reached the highest value of 429.87, and began to converge quickly in the 7th round, and the centralized model performed about the same as the global model; A2C converged the fastest, and the highest cumulative reward can be achieved, the highest value reached 479.99, and the global model reaches the highest value faster than the centralized model; the cumulative reward of the centralized model of TD3 is as good as that of the global model, and the highest value both reached 479.99, but the global model converges in the 3rd round, and the centralized model starts to converge in the 10th round, and the speed of convergence of the global model is 7 rounds earlier than that of the centralized model, which The global model converges faster than the center model, indicating that the global model converges faster; the center model of DDPG is less effective and unstable, which can be seen for the traditional separate DDPG can't achieve to assist the epidemic decision-making, and the global model obtained by using Reinforcement Federated Learning has a relatively better effect and is more stable, and the highest value of the cumulative reward reaches 479.82. Specific experimental results are shown in Tab.~\ref{tab2}:

\begin{table}[htbp]
\caption{Experimental Results of Different Reinforcement Learning Models}
\begin{center}
\begin{tabular}{|c|c|c|c|}
\hline
\multicolumn{2}{|c|}{\textbf{Models}}&\textbf{Cumulative Rewards} & \textbf{Epochs} \\

\hline

\multirow{2}{*}{PPO}    & global & 428.5 & 8  \\\cline{2-4}
                        & center & 438.94  & 8   \\ 
\hline
\multirow{2}{*}{A2C}    & global & 479.99 & 3 \\\cline{2-4}
                        & center & 479.99  & 2   \\ 
\hline
\multirow{2}{*}{TD3}    & global & 479.99 & 3 \\\cline{2-4}
                        & center & 479.99  & 11   \\ 
\hline

\multirow{2}{*}{DDPG}    & global & 479.82 & 2 \\\cline{2-4}
                        & center & 0.0  & none   \\ 
\hline

\end{tabular}
\label{tab2}
\end{center}
\end{table}

\begin{figure*}[htbp]
\centerline{\includegraphics{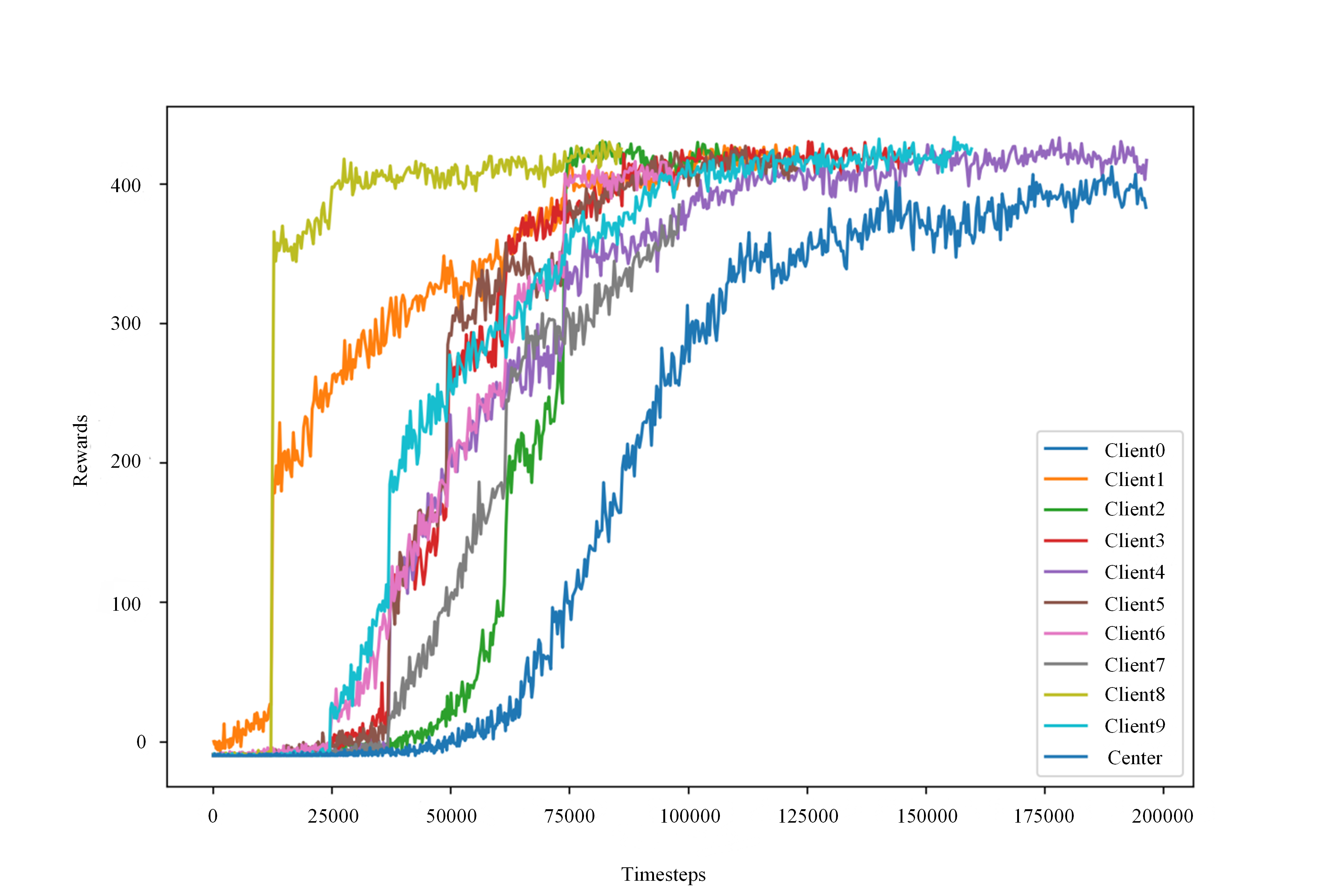}}
\caption{Diagram of Local Training Experimental Results.}
\label{fig4}
\end{figure*}

As shown in Fig.~\ref{fig4}, the training rewards of the client n model after using the reinforcement federated learning method and the training rewards of the model using the centralized training method (center) are shown in Fig. 4. The cumulative rewards of the client n model after using the reinforcement federated learning are higher than the cumulative rewards of the centralized model at the same time step, i.e., the client n local training converges faster and has better effect, which indicates that federated learning can indeed be used for reinforcement learning in epidemic decision-making scenarios and achieve better results, and can help the client intelligences to find the optimal policy quickly while protecting data privacy.

Combining the results of the above experimental comparisons, it can be concluded that the use of reinforcement federated learning can improve both the convergence speed and cumulative rewards of the client's local model, as well as the convergence speed and cumulative rewards of the global model. This is due to the fact that each client under the reinforcement federation learning condition can explore the environment individually at the same time and share the favorable experience gained, which enables the reinforcement learning model to find the optimal policy that is difficult or impossible to find in the central model faster, and thus improves the convergence speed and cumulative rewards of the model.

\section{Conclusion}
In this paper, we propose a method for epidemic decision making using reinforcement federation learning, which can be trained in collaboration with epidemic data from various provinces while protecting privacy. Through experimental comparisons, it is found that using reinforcement federation learning can improve the performance of the decision-making model and can accelerate the convergence speed to achieve faster and better epidemic decision-making, and by comparing different mainstream reinforcement learning models it is found that the use of A2CAvg can achieve the fastest and best performance, and it can be the preferred model for practical use. Future work includes 3 areas:

First, the prediction-based epidemic simulation environment has limited high simulation of reality, and there is a certain difference with reality, which may lead to the epidemic decision-making model obtained by using the environment training can not achieve the expected results in the real world. The success of machine learning and deep learning cannot be achieved without a large amount of data, and if real-world epidemic data can be used to\cite{b37} If we can use real-world epidemic data to train the reinforcement learning model, i.e., offline reinforcement learning\cite{b38,b39}, it will be a perfect solution to the problem of the gap between real and virtual, which will be the next major step.

Second, for federated learning, non-independent and identically distributed data have a large impact on the final effect of the model\cite{b40,b41}. The modeling of outbreak decision-making is not only a good way to improve the performance of the model, but also a good way to improve the performance of the model. Therefore, in order to further verify the performance of reinforced federation learning for epidemic decision-making, it is necessary to observe the effect of reinforced federation learning for epidemic decision-making at this time by using different non-independently and identically distributed epidemic decision-making data for experimental comparison.

Third, to address the impact of non-independently and identically distributed epidemic decision-making data on the reinforcement federated learning model, the use of reinforcement learning to optimize reinforcement federated learning in epidemic decision-making scenarios at different levels of node, migration, and model aggregation can further improve the performance of epidemic decision-making models based on reinforcement federated learning\cite{b42}. Optimization can further improve the performance of the epidemic decision-making model based on reinforcement federated learning. It can also semantically embed and analyze the Chinese text data of the epidemic situation\cite{b43} to assist in epidemic decision-making.

\end{document}